# A PROBABILISTIC MODEL OF COMPOUND NOUNS[1]

MARK LAUER and MARK DRAS

*Microsoft Institute, 65 Epping Road, North Ryde, NSW 2113, Australia*

ABSTRACT

Compound nouns such as *example noun compound* are becoming more common in natural language and pose a number of difficult problems for NLP systems, notably increasing the complexity of parsing. In this paper we develop a probabilistic model for syntactically analysing such compounds. The model predicts compound noun structures based on knowledge of affinities between nouns, which can be acquired from a corpus. Problems inherent in this corpus-based approach are addressed: data sparseness is overcome by the use of semantically motivated word classes and sense ambiguity is explicitly handled in the model. An implementation based on this model is described in Lauer (1994) and correctly parses 77% of the test set.

## 1. Background

### 1.1. Compound Nouns

Levi (1978) provides a good linguistic introduction to the phenomenon of compound nouns. A more computational perspective is afforded by Isabelle (1984). In this paper we shall consider a compound noun to be any sequence of nouns used to refer to something. Thus, *pottery coffee mug*, *hydrogen ion exchange* and *atom bomb missile test exclusion zone perimeter fence* are examples. Such compound nouns pose many difficult problems to automatic language processing systems; this is lucidly argued in Sparck Jones (1983). The main problems fall into three areas: identification of the compound from amongst other text (explored in Arens et al, 1987), the interpretation of underlying semantic relations (explored in Vanderwende, 1993) and the syntactic analysis of the compound. The model presented in this paper will address the last of these.

Traditional grammatical rules used for generating compound nouns are of the form $\overline{N} \rightarrow \overline{N}\ \overline{N}$. This rule, applied recursively, can produce any binary tree over the sequence of nouns. Even assuming that the compound can be identified, this multiplicity of possibilities will give any parser a headache. Nonetheless, this is not a case of overgeneration in the grammar. Examination of a few examples shows that all these possibilities arise: *hydrogen ion exchange* is left branching ( $[[N_1\ N_2]\ N_3]$ ) while *pottery coffee mug* is right branching ($[N_1\ [N_2\ N_3]]$). While context plays an important role in determining the correct analysis, most of the time one of the readings is far preferred simply on the basis of the typical associations between the nouns. Thus it is useful to provide parsers with appropriate preferences according to such associations.

Several systems have been developed which analyse compound nouns. Probably the most sophisticated is that developed in McDonald (1982), in which semantic networks are created for each noun specifying the expected modifiers of that noun. This has the side effect of providing not only a preferred syntactic analysis, but also suggested semantic relationships. Unfortunately, the requirement to provide hand-coded semantic networks for every noun limits the applicability of this approach and the final system is only shown to work on 25 examples. The model proposed in this paper uses knowledge that is provided by the processing of a large corpus in advance. This removes the need for intensive manual coding of semantic preferences and allows for automatic adaptation to specialised domains. We shall assume that the ambiguous compound is presented to the system and that we wish to select one of the possible syntactic analyses (binary branching trees) as the most likely. The general framework adopted is that of probabilistic parsing.

---
[1] To appear in the Proceedings of the Seventh Joint Australian Conference on Artificial Intelligence, World Scientific Publishers, Armidale, NSW, Australia.

*1.2. Specialised Probabilistic Grammars*

Probabilistic parsing systems are based on a model of the language phenomena that they aim to capture. There are several aspects of grammar where achieving a correct analysis requires modeling the effects of individual words. For instance, prepositional phrase attachment depends on the particular preposition. This has given rise to specialised probabilistic grammars, whose coverage is limited to a particular construct of the language. Hindle and Rooth (1993) describe such a system for prepositional phrases. By training on unambiguous examples from a large corpus to acquire lexical preferences, their program achieves almost 80% accuracy on attaching ambiguous prepositional phrases.

Resnik and Hearst (1993) reproduce the system, but add further parameters to the model by taking account of the prepositional object. Unfortunately, even given an enormous corpus, there simply isn't sufficient training data to reliably estimate so many parameters. To circumvent this, Resnik and Hearst propose a technique they call CONCEPTUAL ASSOCIATION. The scheme divides words into categories, assuming that the properties of words are uniform within each category. Probabilities are then estimated for the categories, thus drastically reducing the number of parameters. However, even with this technique, the results do not substantially improve.

Many aspects of prepositional phrases are common to compound nouns. Since these systems have shown that probabilistic models of prepositional phrase attachment can prove successful, a similar approach to compound nouns seems promising. However, it should be noted that compound nouns pose a more difficult problem, since the information provided by the preposition is no longer available.

## 2. A Probabilistic Model of Compound Noun Structure

*2.1. The Need for a Model*

The system proposed in this paper uses counts of occurrences of unambiguous compound nouns to guide the analysis of ambiguous ones. In this respect it is the same as the technique used in Pustejovsky et al (1993). That system brackets compound nouns by searching elsewhere in the corpus for the various possible subcomponents of the compound and then brackets these together. Thus the analysis of *hydrogen ion exchange* depends on whether *hydrogen ion* or *ion exchange* appears elsewhere in the corpus. No evaluation of the technique is provided, but the approach has several disadvantages:

1. It assumes that sufficient subcomponents will appear elsewhere in the corpus
2. It assumes that no spurious subcomponents also appear

**3.** It has no way to indicate the degree of confidence associated with an analysis

**4.** It does not take into account the frequency of words, so that it is biased towards bracketting common words together

The use of a probabilistic model overcomes difficulties 2 through 4. In our model, difficulty 1 is addressed by utilising conceptual association. There are several other advantages in using a well-defined probabilistic model. The model provides a precise denotation for the numerical information used by the program. This is similar to the denotation provided by a formal logic, and allows sound reasoning to be performed. The knowledge used by the system can also be used as the basis of other kinds of inference, for the same reason. Finally, all assumptions of the system are stated explicitly. This is vital when evaluating the applicability of a technique.

*2.2. The Representation*

*Modificational Structure*

According to the general syntax rules of compound nouns, every binary tree with n leaves is a possible parse of a compound noun "$w_1w_2...w_n$" where the $w_i$ are all nouns. Each such parse incorporates a MODIFICATIONAL STRUCTURE. This is defined by assigning one modification relationship for each interior node in the binary tree. In particular, for each interior node, we assert that the rightmost leaf of the left child is a MODIFIER of the rightmost leaf of the right child. This results in (n-1) modification relationships, one for each word except the last. For example, see Figure 1. Unlike the branches of the parse tree, branches in a modificational structure are unordered.

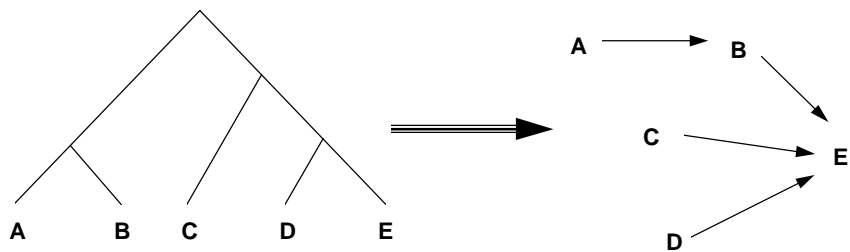

Figure 1 -   Parse trees define modificational structures

This definition of modificational structure follows the general properties of compound noun interpretations, in which the rightmost noun of a given subtree is the head and carries the semantic class of the interpretation of that subtree.[2] Intuitively, the modificational structure represents the meaning structure of the compound, as opposed to the syntactic structure, which can be thought of as being generated from the modificational structure in a productive process. Every word $w_i$ ($i \neq n$) is a modifier of a unique other word further to its right. Hence every modificational structure forms a directed tree with the rightmost word $w_n$ at the root.

---

[2] However, we do not claim that such structures represent the meaning of the compound. They simply create divisions along the lines of such meanings, which are useful for syntactic predictions.

*Equivalence*

Given any directed tree with nodes $w_1, w_2, \ldots w_n$ such that every subtree contains a complete subsequence $w_i, w_{i+1}, \ldots, w_j$ for some $0 < i \le j \le n$, it is easy to show that the tree is a modificational structure derived from exactly one parse tree of the string "$w_1 w_2 \ldots w_n$". It is therefore sufficient to choose the correct structure to determine the correct parse. Henceforth, only the probabilities of modificational structures will be considered, rather than the probabilities of parse trees.

Further, given a modificational structure with nodes $w_1, w_2, \ldots w_n$, every postorder traversal of the tree generates a string "$w_{\theta(1)} w_{\theta(2)} \ldots w_{\theta(n)}$" where $\theta$ is a permutation, which is also a syntactically legal compound noun. In each case (that is, each possible postorder traversal), the modificational structure corresponds to exactly one parse of "$w_{\theta(1)} w_{\theta(2)} \ldots w_{\theta(n)}$" for some $\theta$. See for example Figure 2. Thus, giving any two of the string, the parse or the modificational structure, uniquely defines the third.

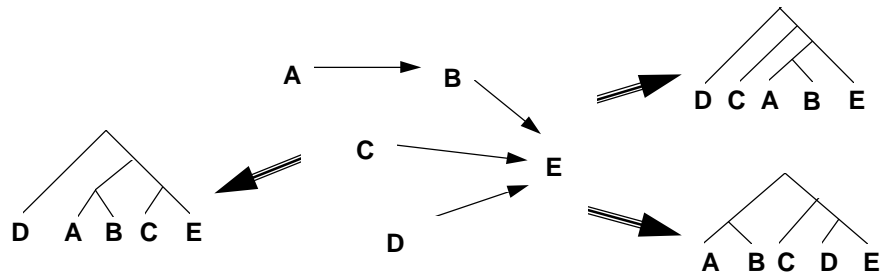

Figure 2 - Modificational structures correpond to one parse
of each of several different strings

*Semantic Classes*

Given the difficulty with sparse data, it is necessary to reduce the number of parameters in the model by considering only conceptual associations. Therefore the model will be based on a set of semantic classes. Probabilities based on these classes will be used on the assumption that all words that realise a given semantic class have roughly the same properties. Let W be the set of all words in compound nouns (each of which will have at least one noun sense). Let S be a set of semantic classes, which are themselves represented by sets of words. That is, every class $s \in S$ is a subset of W. We also assume every word is a member of some semantic class, so that $(\cup_{s \in S} s) = W$. Since words have multiple senses, each word may appear in several semantic classes. Define $cats(w) = \{ s_k | w \in s_k \}$. By the assumption, this is non-empty. Define $ambiguity(w) = | cats(w) |$.

*2.3. Notation*

Each instance of a compound noun is considered an event. We denote the occurrence of a compound whose nouns are $w_1, w_2, \ldots w_n$ in that order by "$w_1 w_2 \ldots w_n$". We also assume that when a word appears in a compound, it is used in a sense that corresponds to one of the semantic classes. Since words are ambiguous, this is not explicit and therefore different sense possibilities must be accounted for by the system. We denote the (unknown) semantic class of a particular instance of a word, $w_i$, by $sense(w_i) \in S$. To allow all word senses in a compound to be considered together, we take $s_1 s_2 \ldots s_n$ to denote the occurrence of a compound "$w_1 w_2 \ldots w_n$" wherein $sense(w_i) = s_i$ for all $0 < i \le n$.

We are interested in modificational structures. Let M(X) denote the set of possible directed trees (with unordered children) whose nodes are elements of the set X. The event m where m ∈ M(W) denotes the occurrence of a compound noun, whose modificational structure is the tree m. When m ∈ M(S), m denotes the event $s_1 s_2 ... s_n$ with the additional information that the modificational structure of the corresponding compound "$w_1 w_2 ... w_n$" is the tree m. Finally, we take $s_i \rightarrow s_j$ to denote the occurrence of a compound whose modificational structure includes the link $w_i$ is a modifier of $w_j$ where *sense*($w_i$) = $s_i$ and *sense*($w_j$) = $s_j$.

*2.4. Assumptions*

*Semantic classes*

To simplify the model, we assume that all semantic classes are equi-probable. That is, the number of occurrences of words used with sense $s_1$ is equal to those used with sense $s_2$, regardless of the actual semantic classes $s_1$, $s_2$. Thus: $P(s_1) = P(s_2)$ and

$$P(s) = \frac{1}{|S|} \quad \text{for all } s \in S \quad (1)$$

This is of particular interest because the assumption makes an assertion about the space of possible meanings, rather than the space of possible syntactic structures as is done in previous models. The relative probability of two semantic classes, $s_1$ and $s_2$, being the sense of a word, w is therefore:

$$\frac{P(s_1|w)}{P(s_2|w)} = \frac{P(w|s_1).P(s_1)/P(w)}{P(w|s_2).P(s_2)/P(w)} = \frac{P(w|s_1)}{P(w|s_2)} \quad \text{using Bayes Rule} \quad (2)$$

*Modificational Structures*

It is an important and novel assumption of the model that all modificational structures involving the same modificational links are equi-probable.[3] In fact, we assume that the probability of a structure is derived from the probabilities of its links by multiplication:

$$P(m) = \prod_{x \in InteriorNodesOf(m)} \left[ \prod_{c \in ChildrenOf(x)} P(c \rightarrow x | \exists z: z \rightarrow x) \right] \quad (3)$$

This assumption differs substantially from other probabilistic grammar models proposed in the past, which typically assume that all parse trees involving the same rewrite rules are equi-probable. Again, our assumption makes an assertion about the distribution of possible meanings rather than about the distribution of possible syntactic structures. The importance of the difference arises because some modificational structures are possible interpretations of several different compound nouns (as is the case in Figure 2), while parse trees are always an analysis of a unique compound.

Intuitively, when a speaker wishes to refer to an entity, she may choose among different possible orderings of the modifiers. For example, suppose an object A has two associated objects B and C, used to identify it. The speaker may use either "$w_B\ w_C\ w_A$" or "$w_C\ w_B\ w_A$". In contrast, if object A is associated with object B, which is associated with object C, the speaker must use "$w_C\ w_B\ w_A$". Therefore, assuming *a priori* equi-probable modificational structures, and given the compound "$w_C\ w_B\ w_A$", the probability of the first structure is half that of the second structure.

---

[3] Since the algorithm only compares the probabilities of structures that generate the same string, only such structures need meet the requirement.

To capture this imbalance between modificational structures, we define the degree of choice available to the generator when given a modificational structure m:

$$choice(m) = \prod_{x \in InteriorNodesOf(m)} NumberOfChildren(x)$$

Note that this measure is independent of the type of the nodes in the tree and applies with $m \in M(W)$ or $m \in M(S)$.

*Generation Process*

Now given a string of elements of X, "$x_1\ x_2\ ...\ x_n$", there is a set of modificational structures, $\Psi_{x_1\ x_2\ ...\ x_n} \subseteq M(X)$, which can generate the string (precisely those for which every subtree contains exactly $x_i, x_{i+1}, ..., x_j$ for some $0 < i \leq j \leq n$). Consider any $m \in \Psi_{x_1\ x_2\ ...\ x_n}$. Since we have no information regarding the generation process, strings that can be generated by m are equi-probable. There are exactly *choice*(m) such strings. Therefore, the probability that m will generate exactly the string "$x_1\ x_2\ ...\ x_n$" is:

$$P("x_1\ x_2\ ...\ x_n"|\ m) = 1/choice(m) \qquad (4)$$

*2.5. Analysis*

Having developed this machinery, we are now in a position to apply it to the problem of analysing compound nouns. The system is based on estimates for the following probabilities:

(i) $P(s_1 \rightarrow s_2 \mid \exists z: z \rightarrow s_2)$ – semantic class $s_1$ modifies $s_2$, given that $s_2$ is modified

(ii) $P(w \mid s_1)$ – word w is used to represent semantic class $s_1$

The choice facing the system is to select some $m \in M(W)$ which is the correct modificational structure for a given compound "$w_1\ w_2\ ...\ w_n$". Thus it must compare the various probabilities $P(m \mid "w_1\ w_2\ ...\ w_n")$ for each $m \in \Psi_{w_1\ w_2\ ...\ w_n}$:

$$P(m \mid "w_1\ w_2\ ...\ w_n") =$$
$$\sum_{s_1 \in cats(w_1),...,s_n \in cats(w_n)} P(m \mid s_1\ s_2\ ...\ s_n) \cdot P(s_1\ s_2\ ...\ s_n \mid "w_1\ w_2\ ...\ w_n")$$

since $s_1 s_2 ... s_n$ is a partition and $P(m \mid s_1 s_2 ... s_n)$ independent of $w_i$'s once $s_i \in cats(w_i)$

$$= \sum_{s_1 \in cats(w_1),...,s_n \in cats(w_n)} \frac{P(s_1\ s_2\ ...\ s_n \mid m) \cdot P(m)}{P(s_1\ s_2\ ...\ s_n)} \cdot P(s_1\ s_2\ ...\ s_n \mid "w_1\ w_2\ ...\ w_n")$$

using Bayes Rule

$$= \sum_{s_1 \in cats(w_1), \ldots, s_n \in cats(w_n)} \left[ \frac{|S|^n \cdot P(m)}{choice(m)} \cdot \prod_{j=1}^{n} P(s_j \mid w_j) \right]$$

since $s_i$ are independent and using (1) and (4) above

$$= \frac{|S|^n}{choice(m)} \sum_{s_1 \in cats(w_1), \ldots, s_n \in cats(w_n)} \left[ P(m) \cdot \prod_{j=1}^{n} P(s_j \mid w_j) \right]$$

since $choice(m)$ independent of $s_i$

where $P(m) = \prod_{x \in InteriorNodesOf(m)} \left[ \prod_{c \in ChildrenOf(x)} P(c \to x \mid \exists z: z \to x) \right]$ from Eq. (3) above.

Thus, given the set of possible analyses, the system can estimate which is the most probable by computing the above function for each.

### 3. Experimental Results

*3.1. Parameter Estimation Process*

Details of an implementation of this strategy are presented in Lauer (1994). Briefly, the 1043 categories of an on-line thesaurus were used as the semantic classes and the parameters $P(s_1 \to s_2 \mid \exists z: z \to s_2)$ were estimated from counts of two word compound nouns in 8 million words from Grolier's encyclopedia as follows:

$$P(s_1 \to s_2 \mid \exists z: z \to s_2) = \sum_{w_1 \in s_1, w_2 \in s_2} \frac{count("w_1 \; w_2")}{ambiguity(w_1) \cdot ambiguity(w_2)}$$

The probabilities $P(w \mid s)$ were estimated as $1/|s|$ which is equivalent to assuming that each word in a thesaurus category is equi-probable.

*3.2. Testing*

A set of 244 three word compounds from the same corpus was hand analysed using the full context. The system was then asked to select one of the two possible analyses, given the probabilistic model above and parameter estimates described. The answers given were correct 77% of the time. For comparison, the baseline correctness achieved by always selecting a left-branching analysis is 67%. This is a promising result in light of the large number of parameters and is similar to the results achieved by prepositional phrase attachment systems (Hindle and Rooth, 1993 and Resnik and Hearst, 1993).

### 4. Conclusion

The problem of syntactic ambiguity challenges NLP systems by creating complexity in the analysis of language. We have argued that specialised probabilistic grammars provide an effective means of acquiring knowledge to address these problems. The use of large corpora as training data for these systems yields practical, accurate and adaptive estimates of useful probabilities. To demonstrate this, we have presented a model of compound nouns and a corresponding algorithm that automatically acquires probabilities for links in the modificational structure. To overcome data sparseness, conceptual association is used which in turn requires the modeling of sense ambiguity. The model has explicitly made a number of assumptions. While most of these would be difficult to validate because of data sparseness, estimates of $P(w \mid s)$ could be derived by expectation maximization (similar to Jelinek et al, 1992). An iterative refinement strategy could also be used (as in Hindle and Rooth, 1993).

Finally, while the goal of this research has been to analyse the syntactic structure of compounds, the problem of determining the underlying semantic relations (as explored in Vanderwende, 1993) remains. While it is not clear how our approach can be extended to perform this task, work such as that reported in Dras and Lauer (1993) suggests that modeling and prediction of such relations is within the reach of this kind of technique.

## 5. Acknowledgments

Invaluable contributions were made by Richard Buckland, Mike Johnson and Robert Dale. Thanks are also due for advice from the anonymous reviewers and Philip Resnik. This work has been supported by the Australian Government via Macquarie University and by the Microsoft Institute.